\documentclass[runningheads]{llncs}
\usepackage[T1]{fontenc}
\usepackage{graphicx}
\usepackage[strings]{underscore}
\usepackage[utf8]{inputenc}
\usepackage{amsmath}
\usepackage{amsfonts}
\usepackage{multicol} 
\usepackage{algorithm}
\usepackage{algpseudocode}

\usepackage{vwcol}

\usepackage{booktabs}
\usepackage{tabularx}
\usepackage{array}

\usepackage{eso-pic}        
\usepackage[dvipsnames]{xcolor}
\usepackage{url}            

\newcommand{\placetextbox}[3]{
  \setbox0=\hbox{#3}
  \AddToShipoutPictureFG*{
    \put(\LenToUnit{#1\paperwidth},\LenToUnit{#2\paperheight}){\vtop{{\null}\makebox[0pt][c]{#3}}}%
  }%
}

\newcommand{\preprintlabel}[2]{%
    \placetextbox{0.5}{0.99}{\colorbox{gray!10}{\textcolor{WildStrawberry}{\textbf{#2}}}}%
    \placetextbox{0.5}{0.97}{\colorbox{gray!10}{\textcolor{WildStrawberry}{\textbf{Author pre-print. The final publication is available at: \url{https://doi.org/#1}.}}}}%
    \placetextbox{0.5}{0.05}{\colorbox{gray!10}{\textcolor{WildStrawberry}{\textbf{Author pre-print. The final publication is available at: \url{https://doi.org/#1}.}}}}%
}

\preprintlabel{10.3390/info16020081}{Published version available!}

\begin{document}

\title{Data Augmentation in Earth Observation: A Diffusion Model Approach}

\author{Tiago Sousa\inst{1}\orcidID{0000-0002-1006-8186} \and
Benoît Ries\inst{1}\orcidID{0000-0002-8680-2797} \and
Nicolas Guelfi\inst{1}\orcidID{0000-0003-0785-3148}}

\authorrunning{T. Sousa et al.}

\institute{Department of Computer Science \\ University of Luxembourg \\ Esch-sur-Alzette, Luxembourg}

\maketitle
\begin{abstract}

The scarcity of high-quality Earth Observation (EO) imagery poses a significant challenge, despite its critical role in enabling precise analysis and informed decision-making across various sectors. This scarcity is primarily due to atmospheric conditions, seasonal variations, and limited geographical coverage, which complicates the application of Artificial Intelligence (AI) in EO. Data augmentation, a widely used technique in AI that involves generating additional data mainly through parameterized image transformations, has been employed to increase the volume and diversity of data. However, this method often falls short in generating sufficient diversity across key semantic axes, adversely affecting the accuracy of EO applications. To address this issue, we propose a novel four-stage approach aimed at improving the diversity of augmented data by integrating diffusion models. Our approach employs meta-prompts for instruction generation, harnesses general-purpose vision-language models for generating rich captions, fine-tunes an Earth Observation diffusion model, and iteratively augments data. We conducted extensive experiments using four different data augmentation techniques, and our approach consistently demonstrated improvements, outperforming the established augmentation methods, revealing its effectiveness in generating semantically rich and diverse EO images.

\end{abstract}

\section{Introduction}

Earth Observation (EO) plays a pivotal role in environmental and geospatial sciences, leveraging remote sensing technology to collect data on Earth's ecosystems, facilitating continuous observations across scales and timeframes \cite{campbellIntroductionRemoteSensing2023}. It is instrumental for monitoring climate change \cite{yangRoleSatelliteRemote2013}, tracking biodiversity \cite{purkisRemoteSensingGlobal2011}, monitoring Sustainable Development Goal (SDG) indicators \cite{andriesUsingDataEarth2022} and assessing ecosystem resilience \cite{sousaModelingPredictingResilience2023}, among others. The integration of Artificial Intelligence (AI) within the EO domain marks a significant advancement, enhancing data processing, analysis, identification and decision making, thus advancing our understanding of Earth's systems \cite{pArtificialIntelligenceApplications2024}. 

However, the application of AI in EO is challenged by data acquisition costs and the limited availability of high-quality and diverse satellite imagery, which are crucial obstacles to address for improving data accessibility, quality, and affordability in advancing AI's use in EO \cite{schmittThereAreNo2023}. Data augmentation has been pivotal in overcoming data acquisition and cost challenges in EO \cite{perezEffectivenessDataAugmentation2017}, focusing on modifying existing datasets through a variety of image transformations \cite{shortenSurveyImageData2019} and by applying generative models \cite{ghaffarDataAugmentationApproaches2019} to expand data volume and diversity. This approach mitigates overfitting and improves AI model accuracy \cite{rebuffiDataAugmentationCan2021}. Yet, traditional data augmentation techniques often falls short in capturing the necessary diversity across key semantic axes of the dataset \cite{yuDeepLearningRemote2017}, crucial for EO. These include detailed attributes like land cover, vegetation types and health, soil materials, water bodies, atmospheric conditions, human-made structures, and temporal changes, all crucial for accurate identification and classification. Failing to encompass this variety can significantly affect the precision and effectiveness of AI-driven EO applications.

This paper addresses the challenges of data scarcity and diversity in EO by introducing a novel data augmentation approach that utilizes diffusion models \cite{rombachHighResolutionImageSynthesis2022}. These models have shown exceptional proficiency in generating synthetic data, achieving state-of-the-art accuracy in fine-tuned tasks \cite{aziziSyntheticDataDiffusion2023}. Our research explores the application of diffusion models in a data augmentation process, specifically designed to enrich the diversity of EO images, aiming to improve the effectiveness of AI applications in this domain.

\section{Related Work}
Data augmentation techniques in AI have been extensively researched \cite{shortenSurveyImageData2019} for their potential to enhance the robustness  and performance of machine learning models \cite{rebuffiDataAugmentationCan2021}, especially in domains where data collection is challenging \cite{bansalSystematicReviewData2022}, such as EO. This section provides an overview of the existing literature, categorizing the techniques into traditional data augmentation techniques, GAN \cite{goodfellowGenerativeAdversarialNetworks2014} based techniques, and emerging strategies involving diffusion models.

\subsection{Traditional Augmentation Techniques}
Traditional data augmentation techniques have primarily focused on the application of parameterized image transformations to enrich datasets. Recently, Hendrycks et al. \cite{hendrycksAugMixSimpleData2020} introduced AugMix, which improves model robustness by mixing various random image transformations. Cubuk et al. \cite{cubukAutoAugmentLearningAugmentation2019} presented AutoAugment, employing a search algorithm to find the best augmentation strategies for specific datasets, significantly boosting model accuracy.

In the context of EO, Abdelhack \cite{abdelhackComparisonDataAugmentation2020} compared several image augmentation techniques for satellite image classification, finding that horizontal and vertical flipping were most effective in achieving high accuracy rates. Illarionova et al. \cite{illarionovaMixChannelAdvancedAugmentation2021} demonstrated an advanced technique where a channel from the original image is replaced with the same channel of another image from the same location but taken on a different date, helping models generalize better to unseen data and outperforming state-of-the-art models in the forest type classification problem. However, its application may not always be feasible in datasets with limited data over the same locations. Yet, traditional data augmentation techniques often fail to adequately diversify and vary the key semantic axes of EO data, which significantly affect the reliability of EO applications, as discussed in \cite{lalithaReviewRemoteSensing2022a}.

\subsection{Augmentation Techniques using GANs}
Subsequent advancements in increasingly sophisticated augmentation techniques, which utilize generative models such as Generative Adversarial Networks (GANs) \cite{goodfellowGenerativeAdversarialNetworks2014}, have shown initial promise in generating realistic images and enhancing the diversity of training datasets beyond image transformations \cite{ghaffarDataAugmentationApproaches2019}. Specifically, GANs have been scaled up in terms of image resolution \cite{ledigPhotorealisticSingleImage2017} and sample quality \cite{brockLargeScaleGAN2019}, which made them more effective than traditional data augmentation techniques. In the domain of EO, GANs have been applied to tackle various challenges and improve the quality of satellite imagery \cite{lalithaReviewRemoteSensing2022a}. For instance, Xiong et al. \cite{xiongImprovedSRGANRemote2020} introduced an improved version of the Super-Resolution GAN \cite{ledigPhotorealisticSingleImage2017} to enhance the spatial resolution of satellite images, improving land cover classification accuracy by 15\% compared to other techniques.

\subsection{Augmentation Techniques using Diffusion Models}
In the rapidly evolving domain of generative models, diffusion models \cite{rombachHighResolutionImageSynthesis2022} have recently gained attention for their capability to generate more detailed and realistic synthetic data with superior quality samples compared to GANs \cite{dhariwalDiffusionModelsBeat2021a}. Furthermore, advancements in diffusion models such as the introduction of classifier free guidance \cite{hoClassifierfreeDiffusionGuidance2021} have enabled text-to-image generation data augmentation \cite{trabuccoEffectiveDataAugmentation2023}, further expanding the range of application areas where realistic generation is crucial. 
However, the potential for data augmentation in EO using diffusion models remains largely underexplored. Recently, Zhao et al. \cite{zhaoLabelFreedomStable2023} presented a novel approach to address the challenge of obtaining high-cost, pixel-level annotations for remote sensing image semantic segmentation. The authors leveraged diffusion models to generate annotation-image pairs from scratch, which achieved competitive accuracy compared to manually annotated data on the dataset, showcasing its potential to significantly reduce the need for laborious annotation processes in remote sensing image analysis. Moreover, Sebaq et al. \cite{sebaqRSDiffRemoteSensing2023b} introduced a novel method for creating high-resolution satellite imagery from text prompts via a two-stage diffusion model. Initially, a diffusion model generates low-resolution images by mapping text to image embeddings in a shared space, ensuring the images capture the desired scenes' essence. Subsequently, another diffusion model refines these images into high-resolution versions with improved detail and visual quality, using the original text prompts. Although the authors did not position their method as a data augmentation technique, their approach demonstrated superior performance on the RSICD dataset \cite{lu2017exploring}.

\section{Earth Observation Data Augmentation} \label{sec-method}

With our approach, we seek to address the limitations of current data augmentation techniques used in EO, which are associated with the lack of sufficient diversity along key semantic axes found in EO satellite or aerial imagery data. We define these key semantic axes as the interpretation and meaning attributed to the visual content of an EO image. These axes typically encompass categories such as land features (e.g., forests, rivers, mountains), natural phenomena (e.g., wildfires, flooding), and human-made structures (e.g., buildings, roads). Additionally, they include the interpretation of environmental or urban landscapes, which might involve recognizing agricultural patterns or urban development.

Augmenting remote sensing image datasets with these axes is crucial for accurately capturing Earth's diverse surfaces. This enhancement is key to improving the robustness and generalizability of machine learning models used for tasks such as land cover classification or object detection, where accuracy is critical. A focal semantic axis is ecosystem resilience, denoting a property of a system or an entity to adapt and recover from failures, often returning to or surpassing a state of normalcy. This is essential for understanding ecosystem responses to challenges like natural disasters, climate change, and human impacts. Incorporating ecosystem resilience into synthetic images enables better modeling of ecosystem impacts and recovery potentials. For example, in disaster management, accurately depicting ecosystem resilience can significantly influence predictions and management strategies for ecological shifts, determining whether an ecosystem quickly rebounds or undergoes long-term degradation.

\begin{figure}
    \centering
    \includegraphics[width=\textwidth]{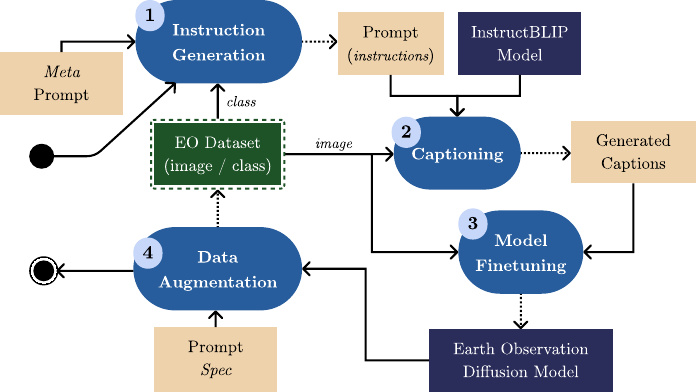}
    \caption{An overview of our four-stage data augmentation process for the generation of synthetic images that accurately depict the intricate interactions and behaviors observed in natural environments.}
    \label{approach-fig}
\end{figure}

Our four-stage data augmentation process, as shown in Figure \ref{approach-fig}, involves the generation of class-parameterized prompts, taking into account the key semantic axes of interest, and derived from an EO training dataset. These prompts are used with a vision-language model to generate captions for the remote sensing images in the dataset. Subsequently, these captions aid in the fine-tuning of a text-to-image diffusion model, which is trained specifically for remote sensing imagery, leveraging the knowledge from the initial dataset. Once fine-tuned, this model serves to augment the dataset by producing additional EO images that are both varied and semantically aligned with the original dataset's context, thereby improving the robustness and diversity of the data available for downstream applications. In the following subsections, we present each stage of our process.

\subsection{Instruction Generation}
The initial stage of our approach involves generating a prompt for each class within the available classes of the EO training dataset. This establishes a one-to-one correspondence between classes and prompts. To facilitate the generation of this set of prompts, we define the following meta-prompt, which is used as input for this stage:
\vspace{-4pt}
\paragraph{Generate a detailed and descriptive caption for the provided remote sensing image, focusing on the specified \texttt{<class>} class. In your description, clearly identify the key characteristics visible in the image. If the image suggests any impact of human activity, natural events, or environmental conditions, elaborate on these.
}\mbox{} \\

\noindent In this meta-prompt, the \texttt{<class>} serves as the parameter to be substituted with the different classes from the training dataset. This parameterization enables the meta-prompt to be adapted to the wide range of classes encountered in EO datasets, offering customized guidance for each class. Furthermore, with this meta-prompt definition and the generated distinctive prompts, we aim to steer the subsequent stage of our approach to focus on the distinct features and implications associated with each \texttt{<class>}, intending to generate detailed, context-aware captions that align with the key semantic axes, previously defined.

\subsection{Captioning}
The second stage involves the application of a vision-language model, specifically to generate captions taking into account the key semantic axes of EO images. To accomplish this, each image from the initial EO dataset is paired with its corresponding class-specific prompt, derived from the previous stage of our approach. These pairs are then input into the InstructBLIP model \cite{daiInstructBLIPGeneralpurposeVisionlanguage2023}, which we chose due to its proven zero-shot performance on large, domain-specific datasets and its ability to extract visual features based on instructions. Such capability is particularly important for the EO domain, where the complexity and diversity of imagery may pose considerable challenges in accurately describing and interpreting visual content for captioning. Despite the domain differences between EO images and the datasets used to train the model, its capability for captioning semantic information in EO images has been demonstrated to be effective \cite{liBLIPBootstrappingLanguageimage2022}.

\begin{figure}[htbp]
  \centering
  \begin{minipage}[c]{0.17\textwidth}
    \includegraphics[width=\textwidth]{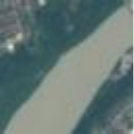}
     \textbf{Class: River}
  \end{minipage}
  \hfill
  \begin{minipage}[c]{0.78\textwidth}
  \textbf{Generated Caption:}
    \textit{The  image is a satellite or aerial view of a river, with a large body of  water visible in the center of the image. The river appears to be quite  long and wide, stretching from one side of the image to the other. In  addition to the river, there are also trees visible in the surrounding  area. No human-made impacts are observed.}
  \end{minipage}
  \caption{An example caption generated from the second stage of our data augmentation process.}
  \label{fig-captioning}
\end{figure}

Building on this foundation, we define a set of configuration parameters for the InstructBLIP model. Consequently, to generate the most probable captions, beam search is performed with a beam width of 5, ensuring a balance between computational efficiency and the quality of generated captions. To avoid overly brief captions, we enforce a minimum token length of 10, while a cap of 256 tokens is implemented to reduce the risk of generating incomplete captions. Finally, a length penalty of -1.0 is strategically set to guide the model towards generating shorter captions, which aligns better with human annotations. At the end of this second stage, a set of captions for the entire EO dataset is generated. In Figure \ref{fig-captioning}, we show an example of how our class-parameterized prompt and configuration parameters allows the InstructBLIP model to produce a detailed, context-aware caption for a "river" image, effectively highlighting its semantic axes and capturing essential visual aspects like land features, natural phenomena, and the environmental landscape.

\vspace{-23pt}
\subsection{Model Fine-tuning}
In the third stage of our proposed approach, we focus on fine-tuning pre-trained diffusion models \cite{rombachHighResolutionImageSynthesis2022} to adapt them to our domain of interest, namely, EO. Diffusion models, a powerful class of generative AI models, are characterized by a forward diffusion process that incrementally corrupts data through the addition of Gaussian noise, using a Markov chain, and a reverse process that aims to denoise the data. The forward process is described by the following normal distribution:
\begin{equation}
q(x_t | x_{t-1}) = \mathcal{N}(x_t; \sqrt{1 - \beta_t} x_{t-1}, \beta_t I),
\end{equation}
where \( x_t \) is the noised image data at step \( t \), \( \beta_t \) the noise variance at step \( t \), and \( I \) the identity matrix. The joint distribution from initial \( x_0 \) to final noisy state \( x_T \) is given by:
\begin{equation}
q(x_{1:T} | x_0) = \prod_{t=1}^T q(x_t | x_{t-1}).
\end{equation}
The reverse process, modeled by a neural network \( p_\theta \) (often a U-Net \cite{ronnebergerUNetConvolutionalNetworks2015}), aims to approximate the true data distribution through optimization of \( \theta \):
\begin{equation}
p_\theta(x_{t-1} | x_t),
\end{equation}
enabling reconstruction of original data from noisy states and genuine sample reconstruction from \( X_T \sim \mathcal{N}(0, I) \).

The application of diffusion models \cite{rombachHighResolutionImageSynthesis2022} has significantly advanced text-to-image generation, producing realistic images guided by natural language descriptions. However, adapting these models for specific domains like EO requires extensive modifications to the model's parameters, resulting in substantial computational challenges. This challenge is further intensified when adapting large, pre-trained models to new domains, with computational constraints often becoming prohibitive. To address this, our approach adopts the Low-Rank Adaptation (LoRA) technique \cite{huLoRALowrankAdaptation2021} for efficient fine-tuning of diffusion models. The advantage of LoRA lies in its capability to fine-tune pre-trained models for new domains using minimal computational overhead by applying a low-rank update to the weight matrices in the transformer's attention mechanism of the diffusion model. Given a pre-trained weight matrix $W \in \mathbb{R}^{n \times m}$, LoRA introduces a low-rank update $\Delta W = AB^T$, where $A \in \mathbb{R}^{n \times d}$ and $B \in \mathbb{R}^{m \times d}$, with $d \ll \min(n, m)$, resulting in $W' = W + AB^T$. This significantly reduces the parameters needing fine-tuning through a process involving the initialization of $A$ and $B$, their adjustment via gradient descent (keeping $W$ static), and the integration of $W'$ into the model.

Therefore, at this stage, we fine-tune a latent text-to-image diffusion model \cite{rombachHighResolutionImageSynthesis2022} using images from the EO dataset along with the previously generated captions. Stable Diffusion V1.5\footnote{https://huggingface.co/runwayml/stable-diffusion-v1-5} weights and LoRA \cite{huLoRALowrankAdaptation2021} are applied for computational efficiency. The training configuration includes mixed precision (\texttt{fp16}) at $512 \times 512$ resolution, random flipping for augmentation, and a batch size of 1. Gradient accumulation is applied over 4 steps to stabilize the training updates, along with 20 epochs, which have been shown to be the optimal duration in our experiments. The initial learning rate is set to $0.0001$ to achieve gradual and stable convergence. Additionally, we cap the gradient norm at 1 to ensure training stability and finally, we apply a constant learning rate scheduler.

The fine-tuned model's ability to produce $512 \times 512$ pixel images provides a strong base for EO applications, especially in Land Use and Land Cover (LULC) classification tasks. Higher resolution images are particularly beneficial in these contexts, as they provide finer granularity and clarity. Such resolution is key for correctly identifying different land cover types and tracking changes over time, thus offering a solid foundation for accurate LULC assessments.

\subsection{Data Augmentation}
In this stage, the trained fine-tuned model is used together with a prompt specification (\textit{Prompt Spec}), to augment the  EO dataset by expanding its volume and diversity. The \textit{Prompt Spec}, a detailed set of instructions where key semantic axes can be defined, are applied to guide the generation of images with the fine-tuned text-to-image diffusion model. Our objective is to use the fine-tuned model to generate an exact number of images for each category in the EO dataset, while allowing the definition of prompts that ensure diversity across critical semantic axes, defined as the interpretation and meaning attributed to the visual content of an EO image.

\section{Experiment}
In this section, we present our experiment designed to compare a variety of data augmentation techniques, including our own, in a classification problem within the EO domain, and to assess the effectiveness of our proposed approach.

\subsection{Dataset}
In our experiment, we use the EuroSAT dataset \cite{helberEuroSATNovelDataset2019}, a deep learning benchmark specifically designed for land use and land cover (LULC) classification. This dataset consists of $27,000$ geo-referenced, low-resolution images \((64 \times 64\) pixels). It is structured into 10 distinct classes: Forest, Annual Crop, Highway, Herbaceous Vegetation, Pasture, Residential, River, Industrial, Permanent Crop, and Sea/Lake, with an average of $2000$-$3000$ images per class. The dataset is partitioned into training (70\%), validation (20\%), and testing (10\%) subsets.

\subsection{Data Augmentation Techniques}

We now present the data augmentation technique strategies used to compare with our approach.

\subsubsection{Baseline} The baseline scenario serves as a control, with no data augmentation beyond resizing images to \(224 \times 224\) pixels. This setup benchmarks our model's performance on the EuroSAT dataset, allowing for evaluation of other augmentation techniques' effectiveness.

\subsubsection{Basic Augmentation} In this strategy, we apply a combination of random resized cropping to \(224 \times 224\) pixels and random horizontal flipping. These transformations are particularly useful in scenarios such as EO, where the orientation of the images can vary, making the model less sensitive to the direction in which the different characteristics are presented.

\subsubsection{Advanced Augmentation} This strategy includes a combination of random horizontal flips, random vertical flips, and random rotations between 0 and 360 degrees, followed by a random resized cropping to \(224 \times 224\) pixels with a scale range from 0.7 to 1.0. With this strategy, we aim to introduce a broader spectrum of variability compared to the basic augmentation technique.

\subsubsection{AutoAugment for Earth Observation} This strategy leverages AutoAugment \cite{cubukAutoAugmentLearningAugmentation2019} with the \textit{ImageNet} policy to apply a wide range of transformations originally optimized for the ImageNet dataset. While the direct applicability of this policy to EO imagery is an area for further research, preliminary findings suggest its effectiveness in enhancing image diversity within EO datasets \cite{liRemoteSensingImage2023}. The rationale behind adopting the \textit{ImageNet} policy in EO contexts lies in its potential to improve model robustness and generalization by exposing the model to a broader spectrum of visual variations.
\label{section-our-augmentation}
\subsubsection{Our Augmentation} We follow our approach presented in Section \ref{sec-method} and resize the images from \(512 \times 512\) pixels (the output resolution of our approach) to \(224 \times 224\) pixels. Additionally, in our \textit{prompt spec}, all prompts use the prefix \textit{"generate a remote sensing satellite image capturing"}. Below is the detailed focus for each prompt, corresponding to each class of the dataset:

\begin{enumerate}
    \item \textbf{AnnualCrop}: Fields of crop fields, arranged in orderly rows with agriculture machines, capturing the agricultural precision.
    \item \textbf{Forest}: A forest landscape as seen from space, highlighting the canopy's texture and diversity, with any visible paths, clearings, or water bodies. In a small area, show signs of deforestation.
    \item \textbf{HerbaceousVegetation}: Areas dominated by herbaceous vegetation, such as meadows or grasslands, showing the texture and color variations of the vegetation
    \item \textbf{Highway}: A major highway as it traverses through various landscapes, including bridges, interchanges, and adjacent urban or rural areas.
    \item \textbf{Industrial}: A large industrial area featuring factories and warehouses, with clear indications of industrial activity such as large parking lots.
    \item \textbf{Pasture}: A lush pasture area from above, with grazing livestock, without trees and a fire close to it.
    \item \textbf{PermanentCrop}: A vineyard, showing permanent crop arrangements, rows of trees or vines, and possibly signs of ongoing maintenance or harvesting.
    \item \textbf{Residential}: A dense residential area with a variety of housing units, surrounded by streets and green spaces.
    \item \textbf{River}: A winding river cutting through diverse landscapes, and adjacent vegetation or urban areas.
    \item \textbf{SeaLake}: A lake or a coastal sea area as seen from above, highlighting the surrounding land, including beaches, docks, or natural vegetation.
\end{enumerate}

\subsection{Model Architecture}
We fine-tune two variants of CLIP \cite{radfordLearningTransferableVisual2021}, namely the ResNet-50 (RN50) and the Vision Transformer B/32 (ViT-B/32) variants, using the EuroSAT dataset. We apply the following parameters for the training pipeline: we utilize the SGD optimizer with a learning rate of $1 \times 10^{-5}$, a momentum of $0.95$, and a small weight decay of $1 \times 10^{-5}$. The optimization process is guided by a cosine annealing learning rate scheduler, with a maximum of 5 iterations and a minimum learning rate set to $1 \times 10^{-8}$. Additionally, we apply cross-entropy loss for both the image and text, where text is defined as the original class of the image. For each data augmentation technique under comparison, the model is trained for 5 epochs, with early stopping applied to prevent overfitting. The model architectures have been implemented using PyTorch and executed on cloud GPUs.

\subsection{Evaluation}
Despite the widespread use of the Fréchet Inception Distance (FID) metric for assessing the diversity of images generated by generative models, we have opted not to employ it in our experiment due to our limited number of synthetic images. FID's reliability increases with the volume of images, and our dataset volume was insufficient for a dependable FID score. Instead, we argue that assessing the impact of generated images on the performance of a model provides a more pertinent measure of the images' utility for our specific EO task. Consequently, we opted to evaluate the impact of the different data augmentation techniques using the top-1 and top-3 accuracy on the test split as our metric. This allows for a straightforward comparison with baseline results and a direct assessment of the generated images practical utility in our specific context.

In an additional evaluation, we assess the effectiveness of two fine-tuned versions of CLIP (RN50 and ViT-B/32), using the same parameters as those outlined in the previous subsection. These models are trained exclusively using images generated through our proposed data augmentation technique and are applied to a zero-shot classification task on the EuroSAT dataset. Subsequently, we compare the performance of these fine-tuned models with the original CLIP results to underscore the impact of our augmentation approach. %
\section{Results}

\begin{figure}[H]
    \centering
    \includegraphics[width=\textwidth]{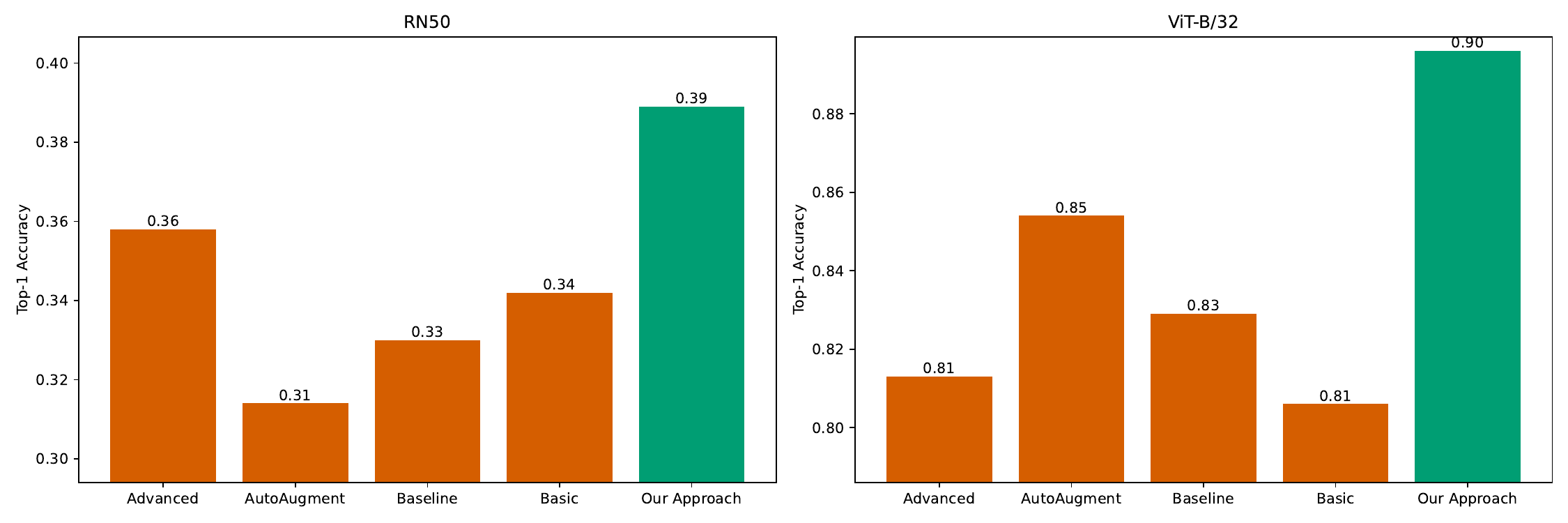}
    \caption{Top-1 Accuracy achieved using the different data augmentation strategies on both CLIP variants.}
    \label{fig-top-1}
\end{figure}

With our experimentation, we demonstrate that our proposed data augmentation approach offers a measurable improvement in model generalization capabilities. This improvement is evident when compared with other previously discussed augmentation methodologies. Notably, with the ResNet50 variant of CLIP, our approach yielded a 3\% increment in top-1 accuracy for the EuroSAT dataset, surpassing the performance of the next best data augmentation technique. The Figure \ref{fig-top-1} illustrates the comparative top-1 accuracy achieved by employing various data augmentation strategies across two distinct architectures, previously presented. Our approach is highlighted, showcasing its superior performance in enhancing model accuracy. For the ResNet50 model, our method achieved a top-1 accuracy of 39\%, which is a significant improvement over the basic and baseline methods, which scored 34\% and 33\%, respectively. In a direct comparison with the AutoAugment strategy, as outlined by Cubuk et al. \cite{cubukAutoAugmentLearningAugmentation2019}, our proposed data augmentation method demonstrates a substantial increase in performance, with a delta of +8\% improvement in top-1 accuracy. Similarly, for the ViT-B/32 variant of our model, our approach achieved a top-1 accuracy of 90\%, outperforming the advanced and AutoAugment strategies, which recorded accuracies of 81\% and 85\%, respectively.

\begin{figure}[H]
    \centering
    \includegraphics[width=\textwidth]{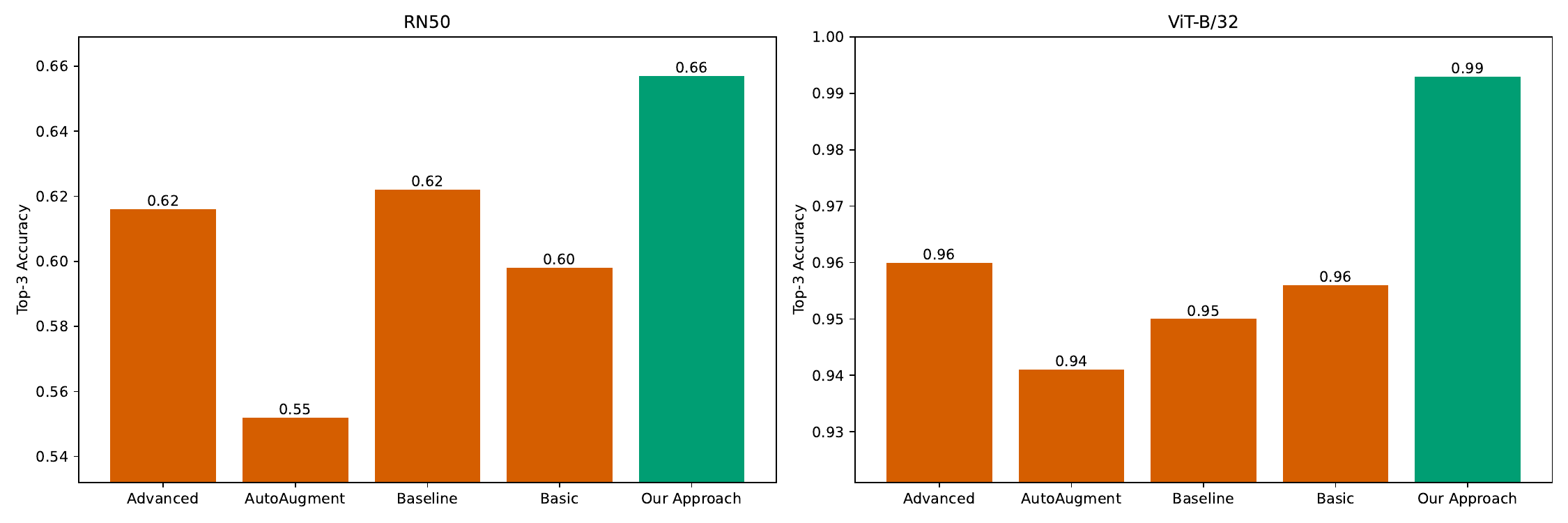}
    \caption{Top-3 Accuracy achieved using the different data augmentation strategies on both CLIP variants.}
    \label{fig-top-3}
\end{figure}

Regarding the top-3 accuracy, depicted in Figure \ref{fig-top-3}, our approach leads to the highest top-3 accuracy, reaching 66\%, which is an improvement over the other strategies. The Advanced and AutoAugment strategies both achieved a top-3 accuracy of 62\%, while the Baseline and Basic strategies resulted in lower accuracies of 62\% and 60\%, respectively. With the ViT-B/32 variant, our approach again demonstrates superior performance with a top-3 accuracy of 99\%, when compared to the other strategies.  These results highlight the effectiveness of our proposed data augmentation and suggest that our method is robust and can be beneficial for improving the performance of models in tasks related to earth observation and remote sensing, thereby confirming the value of our contribution to the domain of EO.

\subsection{Zero-Shot Performance}

\begin{figure}
    \centering
    \includegraphics[width=\textwidth]{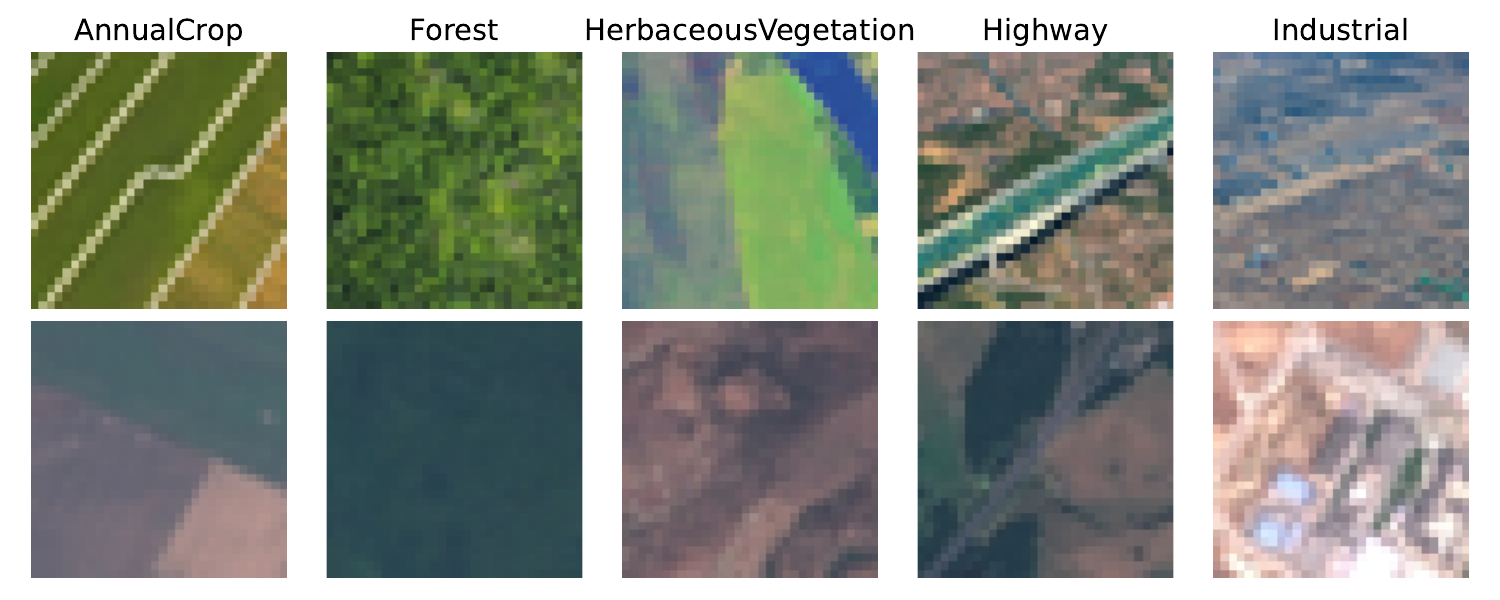}
    \caption{Examples of images generated using our method (top row) compared with images from the corresponding categories in the EuroSAT dataset (bottom row).}
    \label{fig-gen-images}
\end{figure}

In Table \ref{table-zero-shot}, we present the zero-shot performance of models trained exclusively on our generated synthetic images (shown in Figure \ref{fig-gen-images}) and applied to the EuroSAT dataset. The results are quite remarkable, indicating a substantial increase in accuracy compared to the baseline models. These baseline results were obtained from CLIP \cite{radfordLearningTransferableVisual2021}. Specifically, for the CLIP-RN50 model, the zero-shot accuracy achieved using the synthetic training data generated from our proposed data augmentation approach is 58.07\%, which represents an impressive delta of +16.97\% over the baseline accuracy of 41.1\%. Similarly, the CLIP-ViT-B/32 model benefits from our synthetic data, reaching a zero-shot accuracy of 69.23\% on the EuroSAT dataset. This is a significant improvement, with a delta of +19.83\% compared to its baseline accuracy of 49.4\%. These findings illustrate the potential of synthetic data, generated through our proposed data augmentation approach, to significantly improve the zero-shot learning capabilities of models within the domain of earth observation and remote sensing. The marked improvements in accuracy not only affirm the effectiveness of our proposed approach but also demonstrate how it diversifies the earth observation data. This diversity translates into better-performing models that are more adept at handling real-world remote sensing tasks.

\begin{table}[ht]
    \centering
    \caption{Comparison of Zero-Shot Performance on EuroSAT Dataset}
    \label{tab:comparison-zero-shot-acc}
    \begin{tabular}{@{\extracolsep{4pt}}l|c|c|c}
        \toprule
        Model & Accuracy & Ours & Delta \\
        \midrule
        CLIP-RN50 & 41.1\% & 58.07\% & +16.97\% \\
        CLIP-ViT-B/32 & 49.4\% & 69.23\% & +19.83\% \\
        \bottomrule
    \end{tabular}
    \label{table-zero-shot}
\end{table}
\section{Discussion}
During our experimentation, we discovered that current general-purpose vision-language models, including InstructBLIP \cite{daiInstructBLIPGeneralpurposeVisionlanguage2023} and its other variants, are not yet capable to generate accurate and high-quality captions from very detailed meta-prompts. These prompts were designed to elicit captions that capture the essence of the ecosystem depicted in the images, focusing on specific ecosystem types and their intrinsic characteristics such as vegetation, terrain, and climate conditions. Our meta-prompt also requested the model to identify interactions between biodiversity components and the ecosystem services they provide (if any), and to note any observable disturbances along with their impact on ecosystem dynamics. Additionally, we asked the model to employ taxonomy in contextualizing the ecosystem within its broader ecological classification. Despite the comprehensive nature of our meta-prompt, the models frequently produced captions that were not only inaccurate but also suffered from hallucinations, indicating a disconnect between the model's output and the actual content of the images. This experience highlights the current limitations of vision-language models in generating very detailed and domain-specific prompts, which underscores the need for further advancements in the field to meet the demands of accurately captioning EO datasets. %
\section{Conclusion}

In this paper, we present a novel data augmentation approach using generative diffusion models to enrich dataset diversity, showing significant promise in low-data scenarios for enhancing EO classification tasks. Our method not only improves CLIP's zero-shot classification on the EuroSAT dataset but also outperforms existing augmentation strategies in top-1 and top-3 accuracy. Future work could investigate a hybrid strategy combining the generative capabilities of our approach with the variability introduced by advanced augmentation techniques applying geometric transformations, to further improve performance across various levels of data availability. Additionally, evaluating proprietary models like GPT-4V or Claude 3 Opus for generating precise and detailed captions from complex meta-prompts could potentially bridge the gap between current model capabilities and the nuanced requirements of domain-specific prompt generation for EO datasets. 

\bibliographystyle{splncs04}
\bibliography{bib}
\end{document}